\documentclass[journal]{IEEEtran}

\ifCLASSINFOpdf
\else
   \usepackage[dvips]{graphicx}
\fi
\usepackage{url}

\usepackage{subfig}
\usepackage{graphicx}
\usepackage{booktabs}
\usepackage{bm}
\usepackage{amssymb}
\usepackage{amsmath}

\begin{document}

\title{Sorted Convolutional Network for Achieving Continuous Rotational Invariance}

\author{Hanlin Mo and Guoying Zhao, \IEEEmembership{Fellow, IEEE}
\thanks{This work was supported in part by the National Natural Science Foundation of China (Grant No.60873164, 61227802 and 61379082) and the Academy of Finland for Academy Professor project EmotionAI (Grant No.336116).\emph{(Corresponding author: Guoying Zhao.)}}
\thanks{The authors are with the Center for Machine Vision and Signal Analysis, University of Oulu, 90014 Oulu, Finland (e-mail: hanlin.mo@oulu.fi; guoying.zhao@oulu.fi).}}

\markboth{Journal of \LaTeX\ Class Files, Vol. 14, No. 8, August 2015}
{Shell \MakeLowercase{\textit{et al.}}: Bare Demo of IEEEtran.cls for IEEE Journals}
\maketitle

\begin{abstract}
The topic of achieving rotational invariance in convolutional neural networks (CNNs) has gained considerable attention recently, as this invariance is crucial for many computer vision tasks such as image classification and matching. In this letter, we propose a Sorting Convolution (SC) inspired by some hand-crafted features of texture images, which achieves continuous rotational invariance without requiring additional learnable parameters or data augmentation. Further, SC can directly replace the conventional convolution operations in a classic CNN model to achieve its rotational invariance. Based on MNIST-rot dataset, we first analyze the impact of convolutional kernel sizes, different sampling and sorting strategies on SC's rotational invariance, and compare our method with previous rotation-invariant CNN models. Then, we combine SC with VGG, ResNet and DenseNet, and conduct classification experiments on popular texture and remote sensing image datasets. Our results demonstrate that SC achieves the best performance in the aforementioned tasks. 
\end{abstract}


\IEEEpeerreviewmaketitle

\section{Introduction}
\label{section:1}
\IEEEPARstart{F}{eature} extraction is one of the core tasks in computer vision and pattern recognition. An ideal feature should be invariant to various spatial deformations caused by imaging geometry, which ensures that it captures intrinsic information of an image. For many practical applications, such as object recognition, image classification, and patch matching, two-dimensional rotation is the most common spatial transformation, and thus it is essential to achieve rotational invariance of image features in these cases.  

In past decades, numerous hand-crafted features that are invariant to image rotation have been developed \cite{1,2,3,4,5,6}. Since 2012, deep neural networks, especially convolutional neural networks (CNNs), have been proven to be more effective than most hand-crafted features in computer vision tasks. Nonetheless, conventional convolution operations are not rotation-invariant. Actually, even if an image is slightly rotated, CNNs may not be able to recognize it correctly. To address this, a direct approach is to train a CNN with rotated training samples, i.e. data augmentation. However, it has obvious drawbacks, including increasing training time and costs, learning some redundant weights, and further reducing the interpretability of CNNs \cite{7,8}.  

Hence, recent research has aimed to incorporate rotational invariance into convolutional operations and design new network architectures. Based on various methods, such as orientation assignment, polar/log-polar transform, steerable filters, and multi-orientation feature extraction, researchers successively propose Spatial Transformer Network (STN) \cite{9}, Polar Transformer Network \cite{10}, General E(2)-Equivariant Steerable CNN (E(2)-CNN) \cite{11}, Group Equivariant Convolutional Network (G-CNN) \cite{12}, Rotation Equivariant Vector Field Network (RotEqNet) \cite{13}, Harmonic Network (H-Net) \cite{8}, Bessel CNN (B-CNN) \cite{14}, Rotation-Invariant Coordinate CNN (RIC-CNN) \cite{15} and so on \cite{16,17}. 

However, existing rotation-invariant convolution operations have three major limitations: \textbf{1)} Most methods are invariant to specific rotation angles rather than arbitrary angles \cite{12,13,16,17}. Some of them, like RIC-CNN \cite{15}, are only invariant to continuous rotations around image center; \textbf{2)} Some methods require extra trainable parameters and rely on data augmentation when training \cite{9,18,19,20}; \textbf{3)} Many rotation-invariant convolution operations are more complex and not easily replaceable with traditional convolution in common CNN models (like VGG and ResNet) \cite{8,11,12,14}. Moreover, some papers still use data augmentation to train their proposed rotation-invariant CNN models, making it difficult to determine whether the new architectures or only rotated training samples enhanced CNN models' rotational invariance \cite{12,13}.

\begin{figure*}
	\centering
	\subfloat[Square and polar sampling strategies.]
	{\includegraphics[height=20mm,width=62mm]{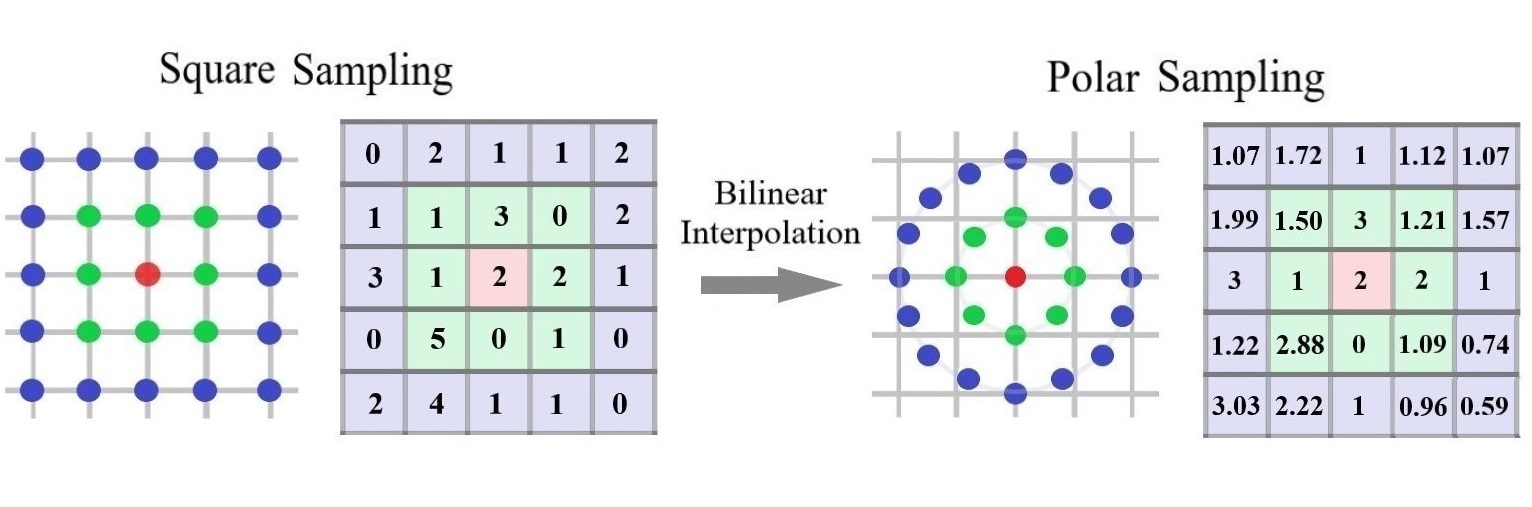}\label{figure:1(a)}\hfill}~~
	\subfloat[Global and ring sorting strategies.]
	{\includegraphics[height=19mm,width=50mm]{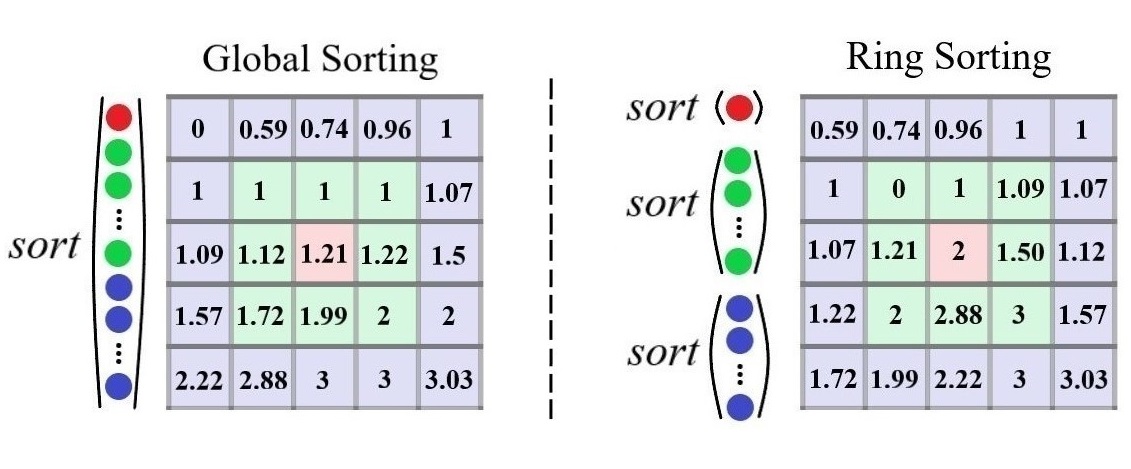}\label{figure:1(b)}\hfill}~~
	\subfloat[Classic and sorted convolution operations.]
	{\includegraphics[height=25mm,width=55mm]{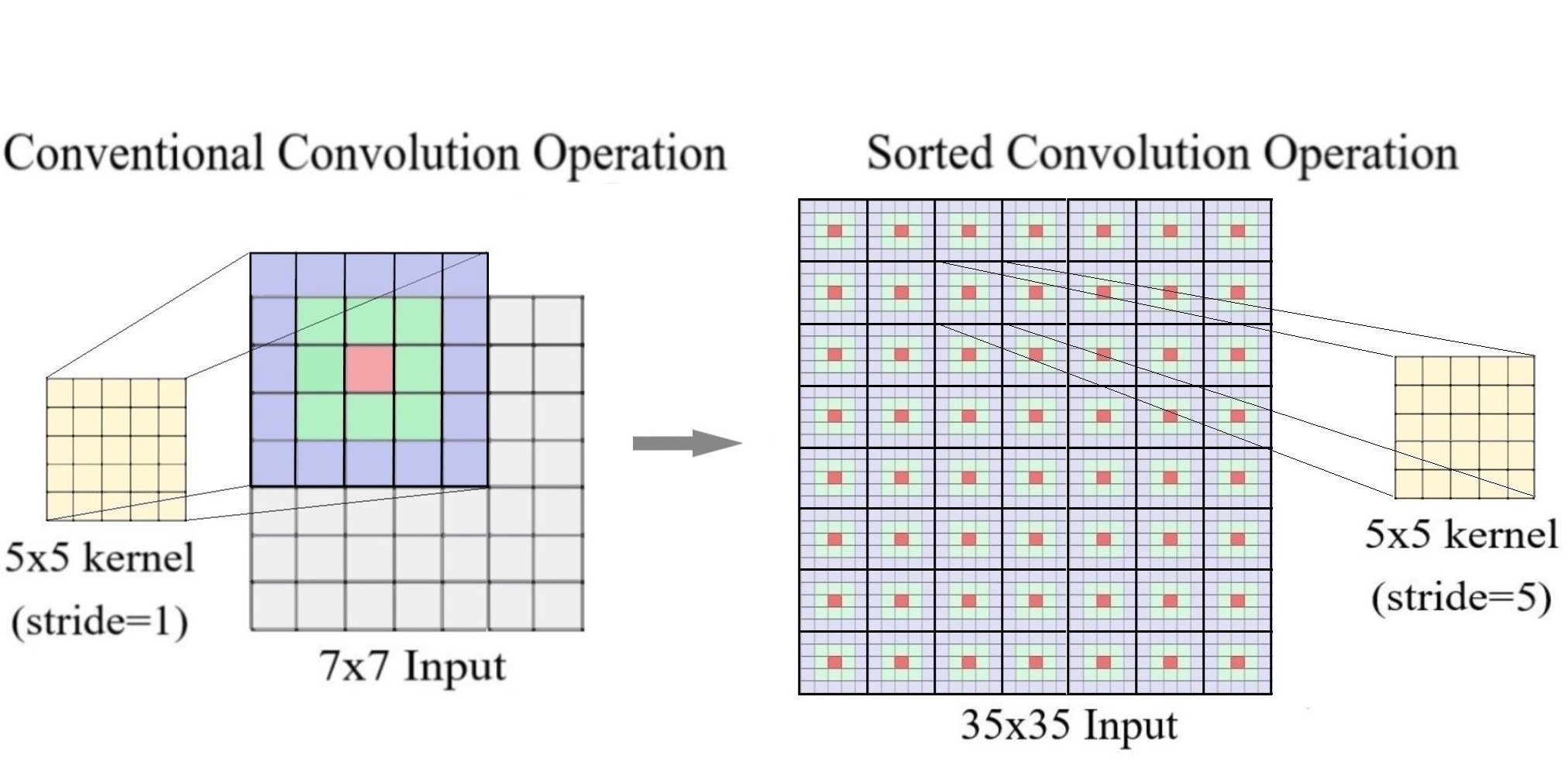}\label{figure:1(c))}\hfill}
	\caption{The implementation details of the proposed sorted convolutional operation.}\label{figure:1}
\end{figure*}

The goal of this letter is to address these limitations in some extend. Our contributions can be summarized as follows: 
\begin{itemize}
	\item Inspired by some hand-crafted features of texture images \cite{21,22,23}, we propose a Sorting Convolution (SC) that achieves continuous rotational invariance without additional learnable parameters. By substituting all standard convolutions in a CNN model with the corresponding SC, we can obtain a Sorting Convolutional Neural Network (SCNN). 
	\item We train SCNN on original MNIST training set without data augmentation, evaluate its performance on MNIST-rot test set, and analyze the impact of convolutional kernel sizes, different sampling methods, and sorting strategies on its rotational invariance. In comparison to previous rotation-invariant CNN models, our SCNN achieves state-of-the-art result. 
	\item We integrate SC into widely used CNN models and perform classification experiments on popular texture and remote sensing image datasets. Our results show that SC significantly increases the classification accuracy of these models, particularly when training data is limited.
\end{itemize}

\section{Methodology}
\label{section:2}

\subsection{Sorted Convolution Operation}
\label{section:2.1}
For an input $F(X)$ with the size of $h \times w$, a conventional convolutional operation $\Phi_{C}$ acting on a given point $X_{0}\in \{1, 2, \cdots, h\}\times\{1, 2, \cdots, w\}$ can be expressed as below
\begin{equation}\label{equ:1}
\Phi_{C}(X_{0},F(X))=\sum_{P\in\mathcal{S}}W(P)\cdot F(X_{0}+P)
\end{equation}
Here, $W$ is a $(2n+1)\times(2n+1)$ learnable kernel, $n$ is a non-negative integer, and $P$ enumerates all points on the square grid $\mathcal{S}=\{-n,-n+1, \cdots, n\}\times\{-n,-n+1, \cdots, n\}$. For example, when $W$ is a $3 \times 3$ kernel, we have $\mathcal{S}=\{(-1,-1),(-1,0), \cdots, (0,1),(1,1)\}$, which contains $9$ points. Our paper only considers odd-sized $W$ because the shift issue occurs in even-sized ones \cite{24}. 

Assuming that $G(Y)$ is a rotated version of $F(X)$, that is, $G(Y) = F(R_{-\theta}Y)$, where $R_{-\theta}$ is a $2\times2$ rotation matrix and $\theta$ represents the rotation angle. Let $Y_{0}$ be the corresponding point of $X_{0}$, then the convolution operation at $Y_{0}$ is
\begin{equation}\label{equ:2}
\Phi_{C}(Y_{0},G(Y))=\sum_{P\in\mathcal{S}}W(P)\cdot G(Y_{0}+P)
\end{equation}
Since $X_{0}=R_{-\theta}Y_{0}$, the following relation can be obtained
\begin{equation}\label{equ:3}
G(Y_{0}+P)=F(R_{-\theta}(Y_{0}+P))=F(X_{0}+R_{-\theta}P)\ne F(X_{0}+P) 
\end{equation}
By substituting (\ref{equ:3}) into (\ref{equ:2}), we can find
\begin{equation}\label{equ:4}
\Phi_{C}(Y_{0},G(Y))\ne \Phi_{C}(X_{0},F(X))
\end{equation}
Thus, the conventional convolutional operation $\Phi_{C}$ is not invariant to two-dimensional rotation. 

Assuming that $(2n+1)\cdot (2n+1)$ points $R_{-\theta}P$ still belong to the square grid $\mathcal{S}$, that is, after rotation, all $R_{-\theta}P$ and $P$ completely overlap. In this case, although $G(Y_{0}+P)\ne F(X_{0}+P)$, we have
\begin{equation}\label{equ:5}
\begin{split}
\{G(Y_{0}+P)\}_{P\in\mathcal{S}}&=\{F(X_{0}+R_{-\theta}P)\}_{P\in\mathcal{S}}\\&=\{F(X_{0}+P)\}_{P\in\mathcal{S}}
\end{split}
\end{equation} 
meaning that the input values used for the convolution operation at points $X_{0}$ and $Y_{0}$ are the same, but with different arrangements. Obviously, if the $(2n+1)\cdot (2n+1)$ values in $\{G(Y_{0}+P)\}_{P\in\mathcal{S}}$ and $\{F(X_{0}+P)\}_{P\in\mathcal{S}}$ are sorted in ascending order separately, the two resulting sorted sequences should be exactly the same. 

If we arrange the sorted sequence in row-major order on the $(2n+1) \times (2n+1)$ square grid $\mathcal{S}$, and represent the new value at point $P$ after sorting as $F^{s}(X_{0}+P)$. Then, the sorted convolution operation can be defined as follows  
\begin{equation}\label{equ:6}
\Phi_{SC}(X_{0},F(X))=\sum_{P\in\mathcal{S}}W(P)\cdot F^{s}(X_{0}+P)
\end{equation}
Since $F^{s}(X_{0}+P)=G^{s}(Y_{0}+P)$ for any $P$, we have
\begin{equation}\label{equ:7}
\Phi_{SC}(Y_{0},G(Y))=\Phi_{SC}(X_{0},F(X))
\end{equation}
indicating that $\Phi_{SC}$ is invariant to arbitrary rotations. Moreover, the sorting operation does not require learning from the training data, so the number of learnable parameters in $\Phi_{SC}$ is the same as in the standard convolutional operation $\Phi_{C}$. 

\subsection{Sampling and Sorting Strategies}
\label{section:2.2}
Formula (\ref{equ:5}) assumes that all $R_{\theta}P$ are still on the $(2n+1)\times (2n+1)$ square grid $\mathcal{S}$, which is only true for discrete convolution when $\theta=k\cdot 90^{\circ}$ ($k$ is an integer). To address this issue, a polar coordinate system centered at $X_{0}$ is established, and $8r$ points are evenly sampled on a circumference with radius $r$ centered at $X_{0}$, where $r=1, 2, \cdots, n$ (see Fig. 1(a)). Bilinear interpolation is used to obtain the values of $F(X)$ at these points, which are then sorted in ascending order and arranged row by row on the square grid $\mathcal{S}$. When the rotation angle $\theta=k\cdot360^{\circ}/(8r)$, the $8r$ points on the circle with radius $r$ coincide before and after rotation. Thus, compared to the square sampling, the polar sampling ensures better validity of the formula (\ref{equ:5}) and rotational invariance of $\Phi_{SC}$.

In addition to the sampling strategy, the sorting strategy is also worth discussing. In fact, the sorting destroys the local structure of $F(X)$ in the $(2n+1)\times (2n+1)$ neighborhood of $X_0$. Previous researchers also used sorting to construct rotation-invariant features for texture images. To preserve the local structure to some extent and improve the discriminability of features, they designed a ring sorting method \cite{21,22}. Unlike the global sorting, the ring sorting separately sorts the $8r$ points in the $r$th square/circular ring about the center point $X_{0}$ and arranges the sorted values in a row-first manner at these $8r$ positions (see Fig. 1(b)). The method preserves some spatial information while ensuring rotational invariance. 

\begin{figure*}
	\centering
	\subfloat[Six SCNN models using square sampling.]
	{\includegraphics[height=28mm,width=53mm]{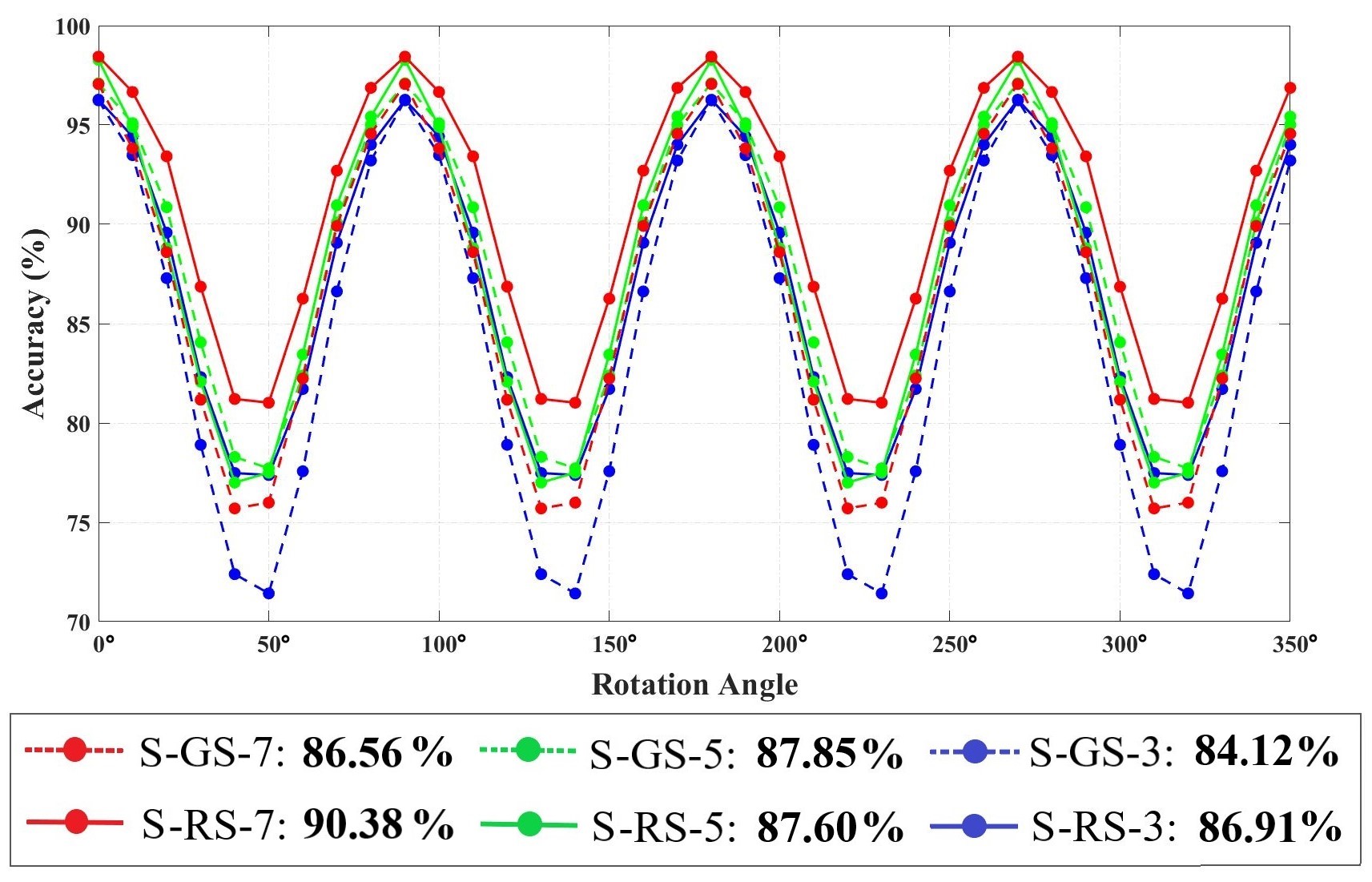}\label{figure:2(a)}\hfill}~~~
	\subfloat[Six SCNN models using polar sampling.]
	{\includegraphics[height=28mm,width=53mm]{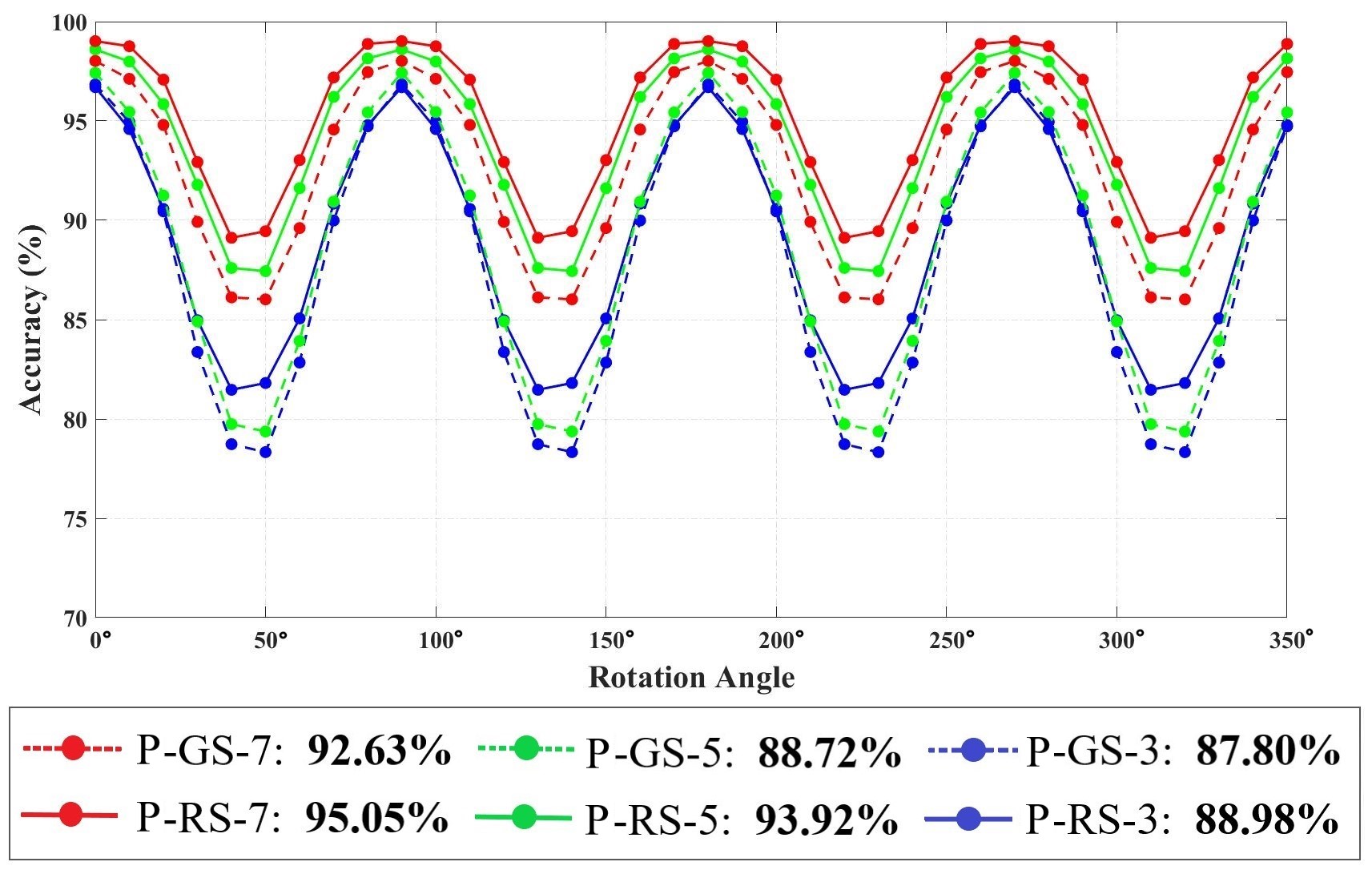}\label{figure:2(b)}\hfill}~~~
	\subfloat[P-RS-7 and some models for comparison.]
	{\includegraphics[height=28mm,width=54mm]{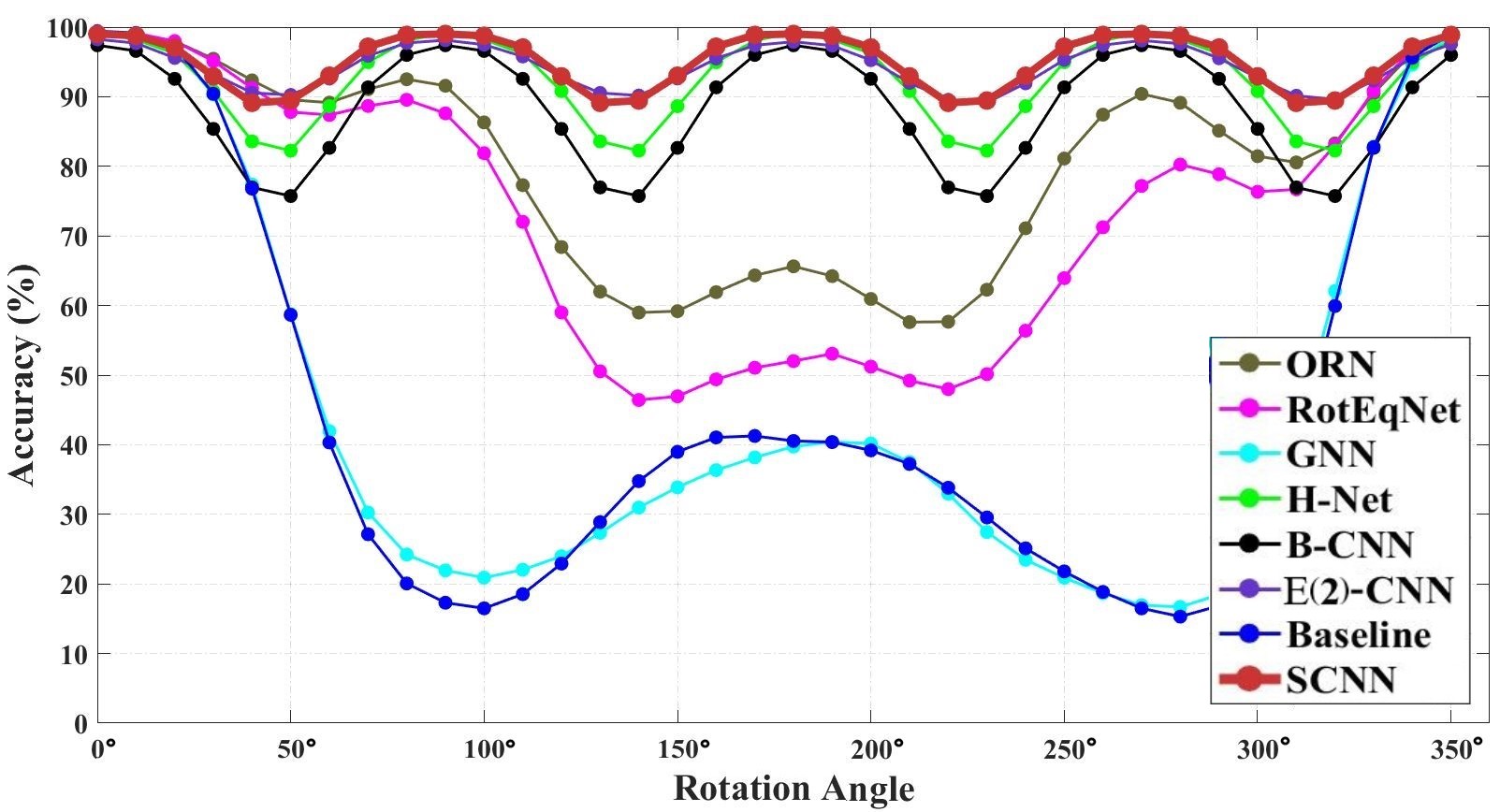}\label{figure:2(c)}\hfill}
	\caption{The classification accuracies from SCNNs and other rotation-invariant CNN models on the MNIST-rot test set.}\label{figure:2}
\end{figure*}

\subsection{The Implementation of Sorted CNN}
\label{section:2.3}
The traditional convolution $\Phi_{C}$ (defined in (\ref{equ:1})) produces an output with the size of $h\times w$ when the stride is set to 1 and padding is performed. However, for the corresponding $\Phi_{SC}$ (defined in (\ref{equ:6})), we first sort the input values within a $(2n+1)\times(2n+1)$ neighborhood for each position $X\in \{1, 2, \cdots, h\}\times\{1, 2, \cdots, w\}$, then concatenate all the sorted neighborhoods to form a new input with size $((2n+1)\cdot h)\times((2n+1)\cdot w)$. Next, we perform a $(2n+1)\times(2n+1)$ convolution on this input with a stride of $(2n+1)$ and no padding is applied. Obviously, the output size of $\Phi_{SC}$ is still $h\times w$, the same as the output of $\Phi_{C}$, while ensuring rotational invariance. Hence, SC and traditional convolutions can be swapped. By replacing all $\Phi_{C}$ in a standard CNN with the corresponding $\Phi_{SC}$, we can create an SCNN. When the input $F(X)$ is rotated by $\theta$ and then input the SCNN, the output of the last SC layer in the SCNN is equal to rotating the corresponding output of the original $F(X)$ by $\theta$. If we use max pooling or average pooling operations to reduce the spatial resolution of the output to $1\times1$, the resulting feature is invariant to arbitrary rotations and can be used as input for other network structures, such as fully connected layers.

\section{Experiments}
\label{section:3}
\subsection{Experiment Setup}
\label{section:3.1}
\textbf{Datasets:} \textbf{MNIST} \cite{24} has 70K $28\times28$ handwritten digit images (0-9), with 60K for training and 10K for testing. 10K training images are randomly selected for validation. Each of test images are rotated from $0^{\circ}$ to $350^{\circ}$ every $10^{\circ}$, resulting in 360K rotated test images. The new test set is called \emph{MNIST-rot} and used to verify the SCNN's rotational invariance. \textbf{Outex\underline{ }TC\underline{ }00012} \cite{25} contains 24 texture classes and 9120 grayscale images of size $128\times128$. For each class, 20 texture surfaces are captured under three lighting conditions ("inca", "t184" and "horizon") as training images. Then, they are captured as test images from 8 different rotation angles ($5^\circ\sim90^\circ$) under "t184" and "horizon" lighting conditions. Thus, the size of the training set is $24 \times 20 \times 3 = 1440$, and the size of the test set is $24 \times 20 \times 2 \times 8 = 7680$. \textbf{NWPU-RESISC45} \cite{26} is a dataset for remote sensing image scene classification. It contains 31500 RGB images of size 256$\times$256 divided into 45 scene classes, each class containing 700 images. We resize all images to 128$\times$128 and randomly select 400 images from each class as training images, with the remaining images used as test images. Due to the arbitrary shooting angles, there are rotational variations present in many classes, such as "airplane", "bridge" and "ground track field".

\textbf{Models and Training Protocol:} We initially design a baseline CNN model with six convolutional layers, having 32, 32, 64, 64, 128, and 128 channels, respectively. We apply $2\times2$ max pooling after the second and fourth layers, and use $7\times7$ average pooling after the final convolutional layer. Then, the feature vector is fed into a fully connected layer with ten units. The kernel size for the last two convolutional layers is $3\times3$, while for the first four layers, the kernel size is the same $K\times K$, where $K\in\{3,5,7\}$. By replacing each of classical convolution operations in the baseline with the corresponding SC, we can obtain a SCNN model. When implementing SC, we have options for square sampling (S) or polar sampling (P), as well as global sorting (GS) or ring-based sorting (RS). This results in 12 different SCNNs $(\{S,P\}\times\{GS,RS\}\times\{3,5,7\})$. For example, P-RS-5 indicates that the first four layers use $5\times5$ SC with polar sampling and ring sorting. We train all these SCNNs on MNIST traning dataset with Adam optimizer, while the initial learning rate is $10^{-4}$, multiplied by $0.8$ every 10 epochs. The number of epochs and the batch size are 100. 

To demonstrate the ease of integrating SC with commonly used CNN models, we select VGG16 \cite{27}, ResNet18 \cite{28}, and DenseNet40 \cite{29} as baseline models. By replacing all traditional convolution operations in these models with SC, we obtain RI-VGG16, RI-ResNet18, and RI-DenseNet40. All of them are trained on Outex\underbar{ }TC\underbar{ }00012 and NWPU-RESISC45, respectively. Again, the Adam optimizer is used, and the training process involves 100 epochs with a batch size of 10. The initial learning rate is set to $10^{-4}$ for VGG16, RI-VGG16, ResNet, and RI-ResNet, while it is $10^{-3}$ for DenseNet and RI-DenseNet. It is reduced by a factor of 0.6 every 10 epochs. 

Our experiments are performed on a Tesla V100 GPU (16G) upon Rocky Linux 8.7 system and PyTorch 2.0.0 framework. All models are trained from scratch without using pretrained parameters or data augmentation. This allows us to directly observe the performance improvement brought by SC. 

\subsection{Results on MNIST-Rot} 
First, we test rotational invariance of 12 SCNN models with varying convolutional kernel sizes, sampling and sorting strategies on the MNIST-rot dataset. This test set contains 36 subsets, each containing 10K samples with the same rotation angle $\theta$ $(0^{\circ}, 10^{\circ}, \cdots, 350^{\circ})$. Fig. 2(a) illustrates the classification accuracy of six SCNNs using square sampling on each subset, while Figure 2(b) displays the accuracy of six SCNNs using polar sampling. Our findings are as follows: \textbf{1)} Polar sampling outperforms square sampling. The accuracy curves in Fig. 2(b) show significant overall improvement compared to Fig. 2(a). For example, S-RS-5 just achieves $87.60\%$ accuracy, whereas P-RS-5 achieves $93.92\%$ on the entire MNIST-rot. This aligns with our theoretical analysis in Section \ref{section:2.2}. \textbf{2)} Ring sorting outperforms global sorting. Notably, P-RS-7 achieves the highest accuracy of $95.05\%$, surpassing S-RS-7's accuracy of $92.63\%$ by $2.42\%$. This is because RS partially preserves spatial information within a convolutional region. \textbf{3)} Using larger convolutional kernel sizes yields better results, especially when combined with ring sorting. For example, the accuracies obtained by P-RS-3, P-RS-5, and P-RS-7 are $88.98\%$, $93.92\%$, and $95.05\%$, respectively.

Fig. 2(c) and Table 1 show the classification accuracies of P-SC-7, its baseline model, and six previous rotation-invariant CNN models on the original MNIST test set and MNIST-rot. Similarly to SCNN, H-Net, B-CNN, and E(2)-CNN also have continous rotational invariance even without data augmentation. In contrast, Oriented Response Network (ORN), RotEqNet, and G-CNN are only invariant to specific rotation angles like multiples of $30^{\circ}$ or $45^{\circ}$. These models are trained using the protocols from their authors. We do not select STN \cite{9}, TI-Pooling \cite{30}, and several methods utilizing rotation-invariant loss functions \cite{31,32} for comparison, because their invariance relies on data augmentation. Our experimental results indicate the following: \textbf{1)} On MNIST-rot, P-SC-7 surpasses the previous state-of-the-art method, E(2)-CNN, by improving the accuracy from $94.37\%$ to $95.05\%$. Additionally, the performance of P-SC-7, H-Net, B-CNN, and E(2)-CNN significantly outperforms ORN, RotEqNet, and G-CNN, highlighting the importance of achieving continuous rotational invariance in CNN models. Furthermore, due to the inability to learn rotational invariance from the training data, even though Baseline and SCNN have an equal number of learnable parameters, Baseline achieves an accuracy of only $44.53\%$. \textbf{2)} On the original MNIST test set, Baseline achieves the best result $(99.43\%)$. Previous research \cite{15} has indicated that rotation-invariant CNNs struggle to distinguish between some digits, like "9" and "6", which contributes to their slightly lower performance on this test set. 

\begin{table}
	\caption{\label{table:1} The classification accuracies on MNIST and MNIST-rot. Bold stands for best results.}
	\centering
	\begin{tabular}{p{1.8cm}p{1.8cm}p{1.8cm}p{1.8cm}}
		\toprule[1.1pt]
		Methods & Input Size & MNIST & MNIST-rot\\
		\toprule[1.1pt]
		ORN\cite{16} & 32$\times$32 & 99.42\% & 80.01\%\\
		RotEqNet\cite{13} & 28$\times$28 & 99.26\% & 73.20\% \\
		G-CNN\cite{12} & 28$\times$28 & 99.27\% & 44.81\% \\
		H-Net\cite{8} & 32$\times$32 & 99.19\% & 92.44\% \\
		B-CNN\cite{14} & 28$\times$28 & 97.40\% & 88.29\% \\
		E(2)-CNN\cite{11} & 29$\times$29 & 98.14\% & 94.37\% \\
		\midrule
		Baseline & 28$\times$28 & \textbf{99.43}\% & 44.53\%\\
		SCNN & 28$\times$28 & 99.04\% & \textbf{95.05\%}\\
		\midrule
	\end{tabular}
\end{table} 


\subsection{Results on Outex\underline{ }TC\underline{ }00012 and NWPU-RESISC45}
We evaluate three commonly used CNN baselines and their corresponding rotation-invariant models on the Outex\underline{ }TC\underline{ }00012 dataset. The rotation-invariant models are obtained by replacing conventional convolutions with SC. The classification accuracy is displayed in the first column of Table \ref{table:2}. Our rotation-invariant CNNs exhibit significantly higher accuracy compared to their baseline counterparts. For example, RI-ResNet18 outperforms ResNet18 by a substantial margin of $34.91\%$. We subsequently reduce the training set size from 1440 to 960 (only training images captured under "inca" and "t184" lighting conditions) and 480 ("inca" lighting condition only). We train six models on these smaller training sets and evaluate their performance on the original test set. The results are shown in the second and third columns of Table \ref{table:2}. Remarkably, even when the training set excludes certain lighting conditions, RI-ResNet18 and RI-DenseNet40 achieve an accuracy of around $98.5\%$. This is because lighting variations do not disrupt the local structure of textures, and the rotation invariance of SC enables it to better extract essential information about local texture structures. Additionally, we conduct classification experiments on the NWPU-RESSC45 dataset, and also reduce the training set size to 1.35K and 0.9K (randomly selecting 300 and 200 images from each category, respectively). Table \ref{table:3} presents the classification accuracies of these models on the test set. Clearly, our rotation-invariant models continue to outperform the corresponding baselines significantly, with a wider gap as the training data decreases. For instance, when the number of training images is reduced from 1.8K to 1.35K and 0.9K, the accuracy difference between RI-ResNet18 and ResNet18 increases from $7.20\%$ to $9.46\%$ and $15.13\%$, respectively.  

\begin{table}
	\caption{\label{table:2} The classification accuracies on Outex\underline{ }TC\underline{ }00012.}
	\centering
	\begin{tabular}{p{2.2cm}p{1.6cm}p{1.6cm}p{1.6cm}}
		\toprule[1.1pt]
		\textbf{Training Data} & \textbf{24}\bm{$\times$}\textbf{40=1440} & \textbf{24}\bm{$\times$}\textbf{40=960} & \textbf{24}\bm{$\times$}\textbf{20=480}\\
		\toprule[1.1pt]
		VGG16  & 60.07\% & 58.13\% & 57.36\%\\
		RI-VGG16 & \textbf{95.99\%} & \textbf{92.90\%} & \textbf{72.28\%}\\
		\midrule
		ResNet18 & 64.79\% & 66.77\% & 63.10\%\\
		RI-ResNet18 & \textbf{99.70\%} & \textbf{99.63\%} & \textbf{98.41\%}\\
		\midrule
		DenseNet40 & 66.02\% & 66.58\% & 60.70\%\\
		RI-DenseNet40 & \textbf{99.47\%} & \textbf{98.62\%} & \textbf{98.53\%}\\
		\midrule
	\end{tabular}
\end{table}

\begin{table}
	\caption{\label{table:3} The classification accuracies on NWPU-RESISC45.}
	\centering
	\begin{tabular}{p{2.2cm}p{1.6cm}p{1.6cm}p{1.6cm}}
		\toprule[1.3pt]
		\textbf{Training Data} & \textbf{45}\bm{$\times$}\textbf{400=1.8K} & \textbf{45}\bm{$\times$}\textbf{300=1.35K} & \textbf{45}\bm{$\times$}\textbf{200=0.9K}\\
		\toprule[1.1pt]
		VGG16  & 71.27\% & 66.32\% & 57.95\%\\
		RI-VGG16 & \textbf{78.53\%} & \textbf{76.61\%} & \textbf{71.39\%}\\
		\midrule
		ResNet18 & 83.18\% & 78.99\% & 70.23\%\\
		RI-ResNet18 & \textbf{90.38\%} & \textbf{88.45\%} & \textbf{85.36\%}\\
		\midrule
		DenseNet40 & 86.83\% & 84.72\% & 80.48\% \\
		RI-DenseNet40 & \textbf{88.35\%} & \textbf{86.96\%} & \textbf{85.93\%}\\
		\midrule
	\end{tabular}
\end{table}

\section{Conclusion}
We develop a Sorting Convolution to achieve continuous rotational invariance in CNNs without additional parameters or data augmentation. Using the MNIST-rot dataset, we analyze the impact of kernel sizes, different sampling and sorting strategies on SC's rotational invariance and compare its performance with other rotation-invariant CNNs. Further, SC can directly replace conventional convolutions in classic CNNs, improving these models' rotational invariance. Thus, we combine SC with commonly used CNN models and conduct classification experiments on popular image datasets. Our results show SC excels in these tasks, especially when training data is limited.

\end{document}